\documentclass{article} % For LaTeX2e
\usepackage{iclr2025_conference,times}
\iclrfinalcopy
\usepackage{amsmath,amsfonts,bm}

\def\eqref#1{equation~\ref{#1}}
\def\1{\bm{1}}

\DeclareMathAlphabet{\mathsfit}{\encodingdefault}{\sfdefault}{m}{sl}
\SetMathAlphabet{\mathsfit}{bold}{\encodingdefault}{\sfdefault}{bx}{n}

\usepackage{url}

\usepackage[utf8]{inputenc} % allow utf-8 input
\usepackage[T1]{fontenc}    % use 8-bit T1 fonts
\usepackage[backref=page]{hyperref}       % hyperlinks
\usepackage{url}            % simple URL typesetting
\usepackage{booktabs}       % professional-quality tables
\usepackage{amsfonts}       % blackboard math symbols
\usepackage{nicefrac}       % compact symbols for 1/2, etc.
\usepackage{microtype}      % microtypography
\usepackage{xcolor}         % colors

\usepackage{float}
\usepackage{graphicx}
\usepackage{amsmath}
\usepackage{amssymb}
\usepackage{amsthm}
\usepackage[ruled]{algorithm2e}
\usepackage{comment}
\usepackage{subcaption}
\theoremstyle{definition}

\usepackage{tablefootnote}

\title{Generating $\pi$-Functional Molecules Using STGG+ with Active Learning}

\author{%
  Alexia Jolicoeur-Martineau \\
  Samsung SAIL Montréal \\
  \texttt{alexia.j@samsung.com} \\
  \And
  Yan Zhang \\
  Samsung SAIL Montréal \\
  \texttt{y2.zhang@samsung.com} \\
  \And
  Boris Knyazev \\
  Samsung SAIL Montréal \\
  \texttt{b.knyazev@samsung.com} \\
  \And
  Aristide Baratin \\
  Samsung SAIL Montréal \\
  \texttt{a.baratin@samsung.com} \\
  \And
  Cheng-Hao Liu \\
  California Institute of Technology \\
  \texttt{chl@caltech.edu} \\
}

\begin{document}

\maketitle

\begin{abstract}
Generating novel molecules with out-of-distribution properties is a major challenge in molecular discovery. While supervised learning methods generate high-quality molecules similar to those in a dataset, they struggle to generalize to out-of-distribution properties. Reinforcement learning can explore new chemical spaces but often conducts 'reward-hacking' and generates non-synthesizable molecules. 

In this work, we address this problem by integrating a state-of-the-art supervised learning method,
STGG+ \citep{stgg+}, in an active learning loop. Our approach iteratively generates, evaluates, and fine-tunes STGG+ to continuously expand its knowledge. We denote this approach STGG+AL.

We apply STGG+AL to the design of organic $\pi$-functional materials, specifically two challenging tasks: 
1) generating highly absorptive molecules characterized by high oscillator strength and 2) designing absorptive molecules with reasonable oscillator strength in the near-infrared (NIR) range. The generated molecules are validated and rationalized \textit{in-silico} with time-dependent density functional theory. Our results demonstrate that our method is highly effective in generating novel molecules with high oscillator strength, contrary to existing methods such as reinforcement learning (RL) methods. 

We open-source our active-learning code along with our Conjugated-xTB dataset containing 2.9 million $\pi$-conjugated molecules and the function for approximating the oscillator strength and absorption wavelength (based on sTDA-xTB). \\ \textbf{Code:} \url{https://github.com/SamsungSAILMontreal/STGG-AL}.

\end{abstract}

\begin{figure}[htbp]
    \renewcommand{\figurename}{Task}
    \vspace{-10pt}
    \centering
    \begin{minipage}{0.48\textwidth} 
        \centering 
        \includegraphics[width=0.75\linewidth]{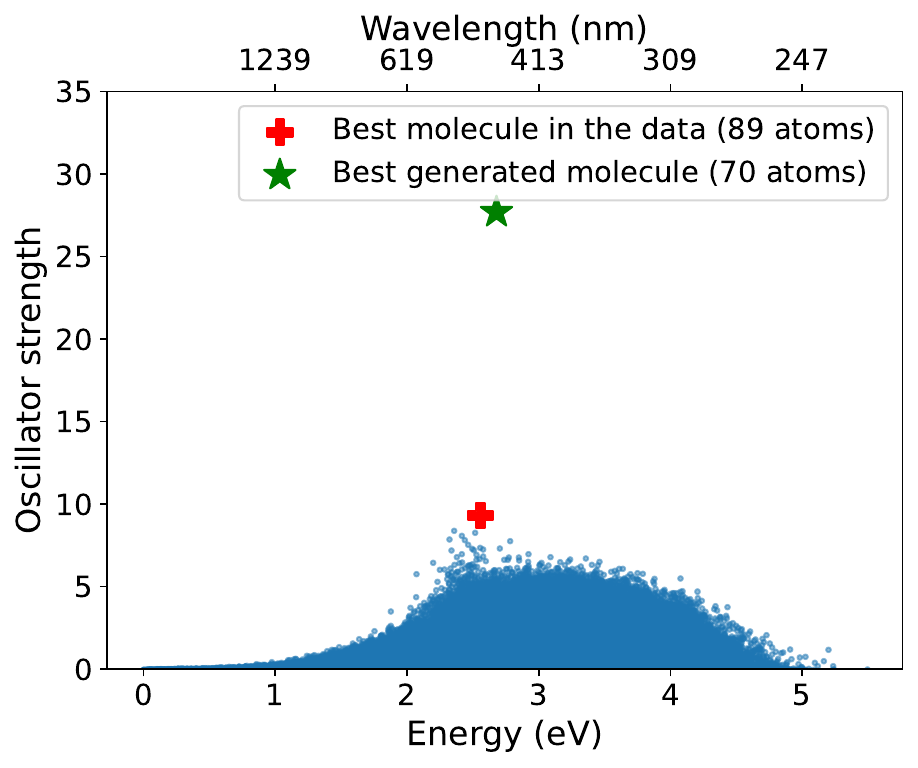}
        \caption{Maximizing $f_\text{osc}$. STGG+ with active learning generates strong out-of-distribution (OOD) molecules.} 
        \label{fig:Normal_fig_train}
    \end{minipage}\hfill 
    \begin{minipage}{0.48\textwidth} 
        \includegraphics[width=0.75\linewidth]{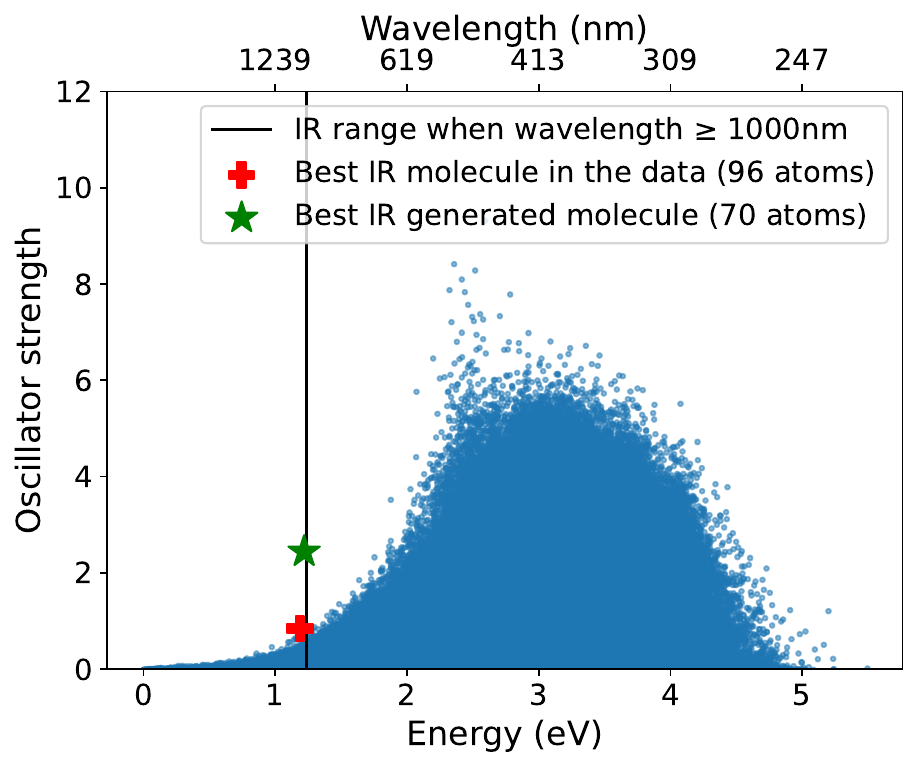} 
        \caption{Maximizing $f_\text{osc}$ in the short-wave infrared range. STGG+ with active learning generates strong OOD molecules.} 
        \label{fig:IR_fig_train}
    \end{minipage}
    \vspace{-10pt}
\end{figure}

\section{Introduction}

Generating novel organic molecules with desirable, previously unseen optoelectronic properties holds transformative potential across many applications, from display technology to wearable electronics to biomedical imaging \citep{review1,review2}. Central to this pursuit are $\pi$-conjugated functional molecules, where their $\pi$-delocalized electrons enable functionalities such as in organic light-emitting diodes (OLED) and short-wave infrared (SWIR) absorbers. Traditional approaches to molecular design, however, face the persistent challenge of systematically exploring uncharted regions in the chemical space to identify out-of-distribution properties while remaining chemically reasonable. 

Supervised learning methods typically address this problem by modeling the distribution of a given dataset, but extrapolating beyond the training set (i.e., out-of-distribution generalization) is difficult. Effective molecular generation requires generative models to capture meaningful patterns (e.g., chemical rules) that enable generalization.

Unsupervised methods such as Reinforcement learning (RL) \citep{reinvent,reinvent4,release} do not rely on datasets and instead generate molecules and evaluate them progressively. Although powerful, with imperfect reward models in chemistry, RL can exploit the reward function and generate chemically non-viable molecules unless carefully regularized. 

Active learning \citep{settles2009active} holds promise in combing both worlds (supervised and unsupervised) by training a model using supervised learning and then iteratively generating new molecules, labeling them, and continuing training the model with them \citep{merchant2023scaling, korablyov2024generative, kyro2024chemspaceal, antoniuk2025active}. This approach allows the joint sampling from a strong base model and the reward model. It is more aligned with how humans learn: chemists learn about molecules from existing literature, then they generate new molecules, test them, and then rebuild their own priors about which aspects of the molecule lead to better properties.

STGG+ is an autoregressive generative model that uses spanning tree-based graph generation and is trained in a supervised manner with strong in-distribution and out-of-distribution capabilities \citep{stgg+, stgg}. In this work, we propose to combine STGG+ with active learning to design $\pi$-conjugated molecules with out-of-distribution optoelectronic properties, a challenging problem which current RL methods struggle with. 

We explore two proof-of-concept yet application-oriented challenges. Specifically, we design: 
\begin{itemize}
    \item Molecules with exceptionally high oscillator strength ($f_\text{osc}$), which correlates with efficient photo-absorption/emission, relevant for designing OLED materials~\citep{abroshan2022machine}. 
    \item Molecules with high $f_\text{osc}$ and strong absorption in targeted spectral ranges, particularly in NIR for potential biomedical imaging applications~\citep{privitera2023shortwave}.
\end{itemize}

We constructed a computational dataset of 2.9 million $\pi$-conjugated molecules and pre-trained STGG+ on it, followed by active learning. Our results show that STGG+ combined with active learning can progressively move toward higher $f_\text{osc}$ molecules much more effectively than baseline methods such as genetic algorithms and reinforcement learning. Active learning required only 30,000 additional data points to discover candidates with $f_\text{osc}$ of 27.7, compared to a maximum of 9.3 found through traditional virtual screening. This is not only a great improvement, but also a significant speedup compared to virtual screening considering that reward evaluation requires expensive simulation. Furthermore, molecules generated by RL tend to have issues with chemical validity or synthesizability (e.g. exotic ring structures), while STGG+ generates chemically sound molecules by design. We validated the top-1 generated samples using time-dependent (TD) density functional theory (DFT), which explain the new scaffolds.

\section{Method}

In this work, we seek to maximize $f_\text{osc}$ while maintaining chemical reasonableness and some additional constraints. More generally, assume that we aim to generate molecules with out-of-distribution properties not seen in the dataset. Some properties need to be maximized ($\Lambda$), while others need to be constrained within some range of values ($\Omega$). Our approach is described below.

First, we pre-train STGG+ on some dataset(s) with molecules similar to those desired conditioned on their properties. 
We fix the range of properties for the constraints $\Omega \sim \mathcal{U}(\Omega_{min}, \Omega_{max})$ and initialize the properties to be maximized $\Lambda \sim \mathcal{U}(\Lambda_{min}$, $\Lambda_{max}$) to be around (slightly-lower and slightly-above) the maximum values found in the dataset.

Then, we iterate through $N$ steps of active learning: 
\begin{enumerate}
    \item Generate $k$ molecules from STGG+ conditioned on the sampled properties. \\ $\Lambda \sim \mathcal{U}(\Lambda_{min}, \Lambda_{max})$ and $\Omega \sim \mathcal{U}(\Omega_{min}, \Omega_{max})$
    \item Remove invalid/duplicated molecules.
    \item Evaluate the generated molecules to determine their properties using the pipeline in Sec.~\ref{sec:dataset}. 
    \item Update the range of $\Lambda_{min}$, and $\Lambda_{max}$ to be respectively the top-1 and top-100 properties to maximize ($\Lambda$) of the generated molecules (slowly expanding the Pareto frontier).
    \item Fine-tune STGG+ on the generated molecules conditional on their properties. 
\end{enumerate} 

\section{Conjugated-xTB dataset}
\label{sec:dataset}

We present a dataset containing 2.9 million $\pi$-conjugated organic molecules. The molecules are constructed by sampling a set of 181 hand-curated $\pi$-conjugated molecular fragments and connecting them at allowed atomic indices. All molecules have between 4-8 fragments and a maximum of 100 heavy atoms. We did not consider solubility, but alkyl chains can be readily appended to the building blocks. The 181 fragments represent common, synthesizable building blocks which we classify into electronic donors, acceptors, and 'neutral' connecting bridges. While the dataset is not optimized for synthesizability, the combinations of these building blocks are expected to represent most motifs of optoelectronically-active molecules. On average, each fragment has 2.77 connections; for 4-8 fragments, the total number of molecules that can be constructed using these fragments (without atom limitation) are respectively $~\sim 6\times 10^{10}, 3\times 10^{13}, 2\times 10^{16}, 8\times 10^{18},~ \text{and}~ 4\times 10^{21}$. 

For each sampled molecule, we generate 32 conformations using ETKDG as implemented in RDKit~\citep{riniker2015better}, and these geometries are optimized by MMFF94 forcefield~\citep{halgren1996merck}. The lowest-energy conformer is selected for further geometry optimization using the semiempirical quantum chemistry method GFN2-xTB~\citep{bannwarth2019gfn2}. We then approximate the optical properties of this conformer using sTDA-xTB \citep{grimme2016ultra}, which calculates the vertical absorption energy and the corresponding $f_\text{osc}$~\citep{grimme2016ultra}. The dataset contains 2.9 millions rows and 3 columns (SMILES, $f_\text{osc}$, absorption wavelength (in nm)). We open-sourced the full dataset. 

\section{Experiments}

We run experiments on two problems. First, we seek to maximize $f_\text{osc}$, which is correlated to the intensity of absorption/emission processes. Second, we aim to maximize $f_\text{osc}$ in the short-wave infrared absorption range (absorption wavelength $\lambda_{abs} \ge 1000$ nm), which is crucial for biomedical imaging as tissues exhibit relatively low absorption and scattering in NIR, allowing for deeper penetration of light \citep{wilson2015review, privitera2023shortwave}.

To increase the chances of synthesizability, we also impose a maximum ring size of 6 and a maximum number of heavy atoms of 70. STGG+ also imposes proper valency \citep{stgg} by its design.

STGG+ is first pre-trained on the Conjugated-xTB dataset for 5 epochs. Then, active learning is applied. To maximize diversity, we uniformly sample a temperature between 0.7 and 1.1 and classifier-free guidance \citep{ho2022classifier} between 0.5 and 1.5. We generate 2000 molecules per active learning step, and they are trimmed down (removing duplicates and invalid molecules). Fine-tuning is done for 100 epochs on the last batch of 2000 generated molecules.  The other hyperparameters follow the default ones by \citet{stgg+}. More training and architectures details can be found in Appendix \ref{app:1}-\ref{app:2}.

We compare STGG+ to two strong baselines (as mentioned by \citet{tripp2023genetic}): GraphGA \citep{jensen2019graph, brown2019guacamol}, and REINVENT4 \citep{reinvent4}, version 4 of the popular REIVENT \citep{reinvent}. For REINVENT4, we use the default settings. For GraphGA, we used the good choice of hyperparameters suggested by \citet{tripp2023genetic} consisting of 5 new candidates per generation and running as many generations as possible. The baseline methods are given the top 100 molecules from the Conjugated-xTB dataset as initial starting points.
We run each algorithm long enough to reach around 30K evaluations. Since RL can be quite noisy compared to supervised learning, we do 3 runs of the RL baselines using 3 different seeds. See Appendix \ref{app:rl} for more details on the RL baselines.

The results are described in the subsections below. We also describe the top-10 molecules made by STGG+ in Appendix \ref{app:3} and show the top-1 molecules made by all methods in Appendix \ref{app:4}.

The geometries of top-1 molecules are selected to be further optimized in DFT with the B3LYP hybrid functional and def2-SVP basis set. Single-point TD-DFT calculations are then computed to cross-check with sTDA-xTB vertical absorption energies/oscillator strength.

\subsection{Task 1: Maximizing the oscillator strength}

\begin{figure}[h]
    \centering 
    \includegraphics[width=0.5\linewidth]{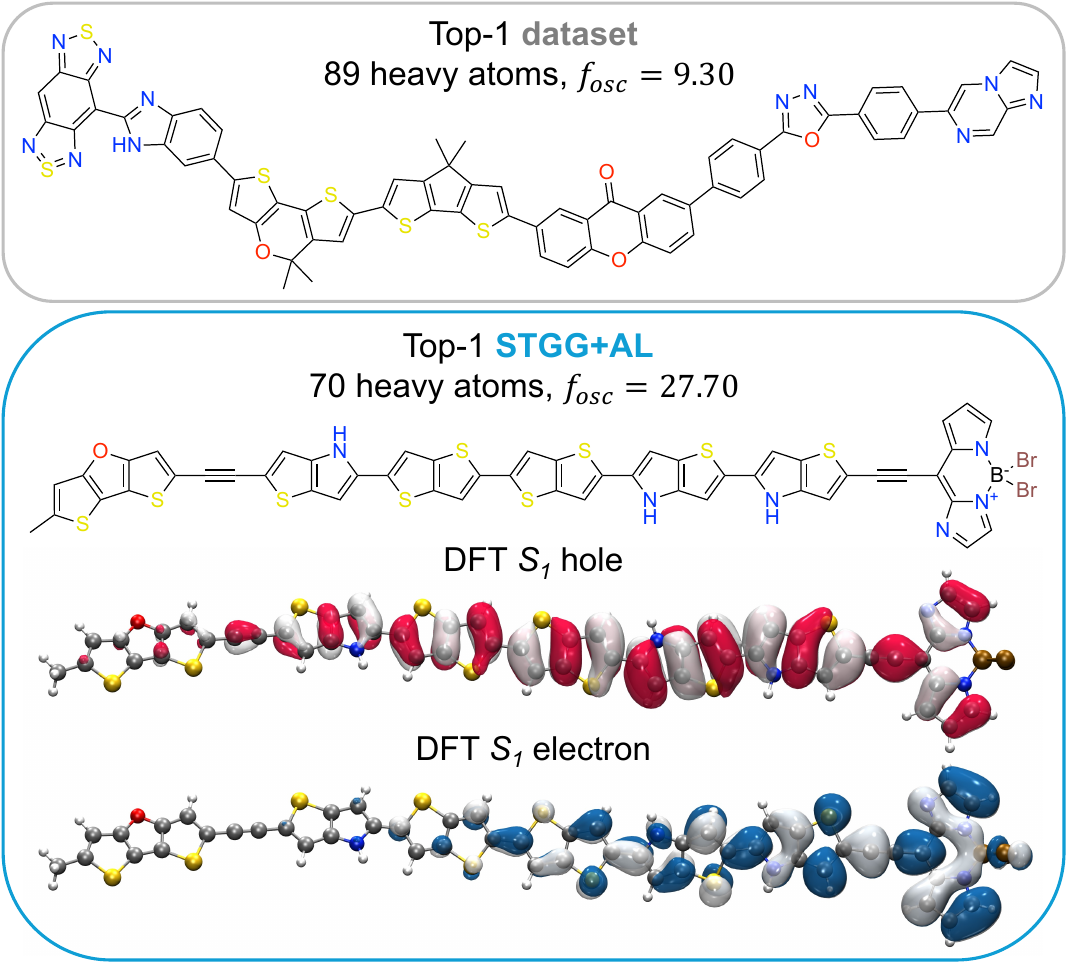}
    \caption{Case study of the top-1 molecule with the highest $f_\text{osc}$.}
    \label{fig:fosc}
    %\phantomsubcaption\label{fig:data_best}
    %\phantomsubcaption\label{fig:fosc}
    \end{figure}

\begin{figure}[h]
    \centering
    \begin{minipage}{0.4\textwidth} 
        \centering 
        \includegraphics[width=\linewidth]{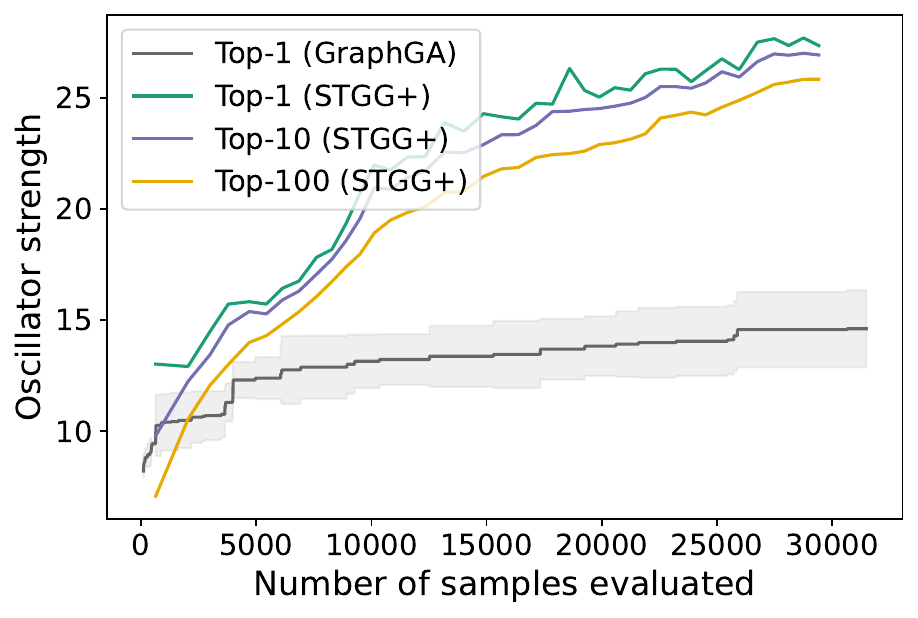}
    \end{minipage}%\hfill 
    \begin{minipage}{0.4\textwidth} 
        \centering
        \includegraphics[width=\linewidth]{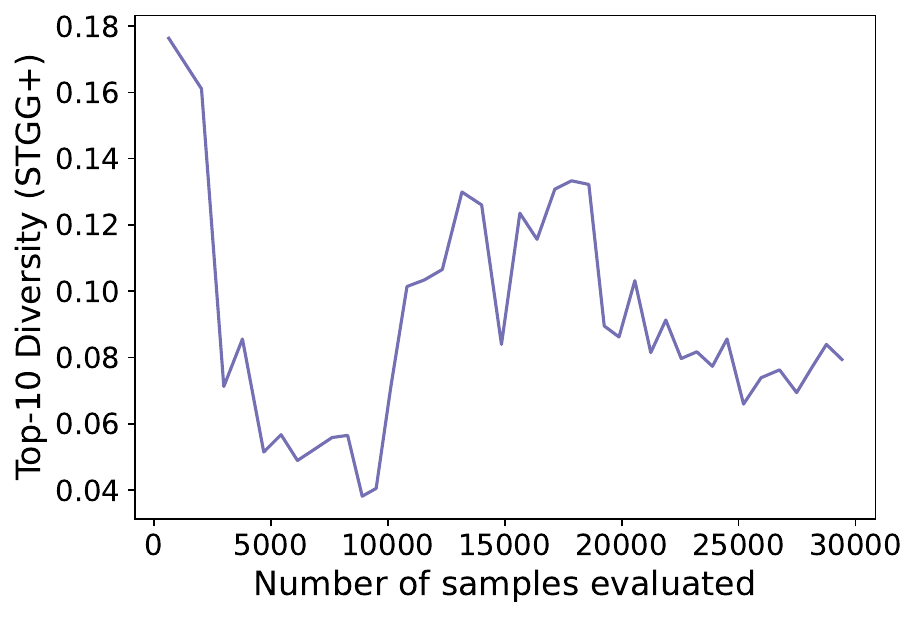} 
    \end{minipage}
    \caption{Maximizing $f_\text{osc}$ using active learning with constraints: max 70 heavy atoms, max ring-size of 6. STGG+ (top-1, top-10, top-100; from a single run) vs GraphGA  (top-1; average and 95\% confidence interval over 3 runs).}
    \label{fig:stgg_fosc1}
\end{figure}

Figure \ref{fig:fosc} shows the molecule with the highest $f_\text{osc}$ from the Conjugated-xTB dataset, and the molecule with the highest $f_\text{osc}$ generated by STGG+ with active learning given the constraints ($\leq$ 70 atoms, max ring size of 6). Figure \ref{fig:stgg_fosc1} shows the progress over time. STGG+ learns to sample rigid and planar molecules, which can have high orbital overlap and hence high $f_\text{osc}$. We see that mini-batch diversity initially drops down at around 5K Oracle calls, then moves back up at 10K Oracle calls and stabilizes. 

TD-DFT calculations show a high $f_\text{osc} = 3.79$\footnote{sTDA-xTB shows a $f_\text{osc}=27.70$. We note that $f_\text{osc}$ from different quantum chemical methods may not be directly comparable.}. The natural transition orbital (NTO) of the first excited state confirms large hole/electron overlap over the rigid backbone.

\subsection{Task 2: Maximizing the oscillator strength in the short-wave infrared range of absorption}

Figure \ref{fig:fosc} shows the highest $f_\text{osc}$ molecule in the short-wave infrared range from the Conjugated-xTB dataset, and the highest $f_\text{osc}$ molecule in the NIR range generated by STGG+ with active learning given the constraints ($\leq$ 70 atoms, max ring size of 6). Here, STGG+ learns to generate different scaffolds of charge-transfer species, which can explain their lower absorption energy. We see that mini-batch diversity drops slowly over time, showing convergence toward some regions of the molecular space. Improvement in oscillator strength over time is somewhat linear, showing that STGG+AL could improve further if more Oracle calls were given.

The case study in Figure \ref{fig:fosc} demonstrates a semi-symmetric scaffold that does not have highly electron-rich nor -deficient fragments, hence it is not expected to bear low energy transitions. 

TD-DFT calculations confirms the NIR absorption wavelength found in sTDA-xTB, where the $S_0\leftrightarrow S_1$ transition is at 973 nm, with a small but not negligible $f_\text{osc}=0.3$. NTO analysis showcases a charge-transfer behavior with a small orbital overlap.

\begin{figure}[h]
    \centering 
    \includegraphics[width=0.6\linewidth]{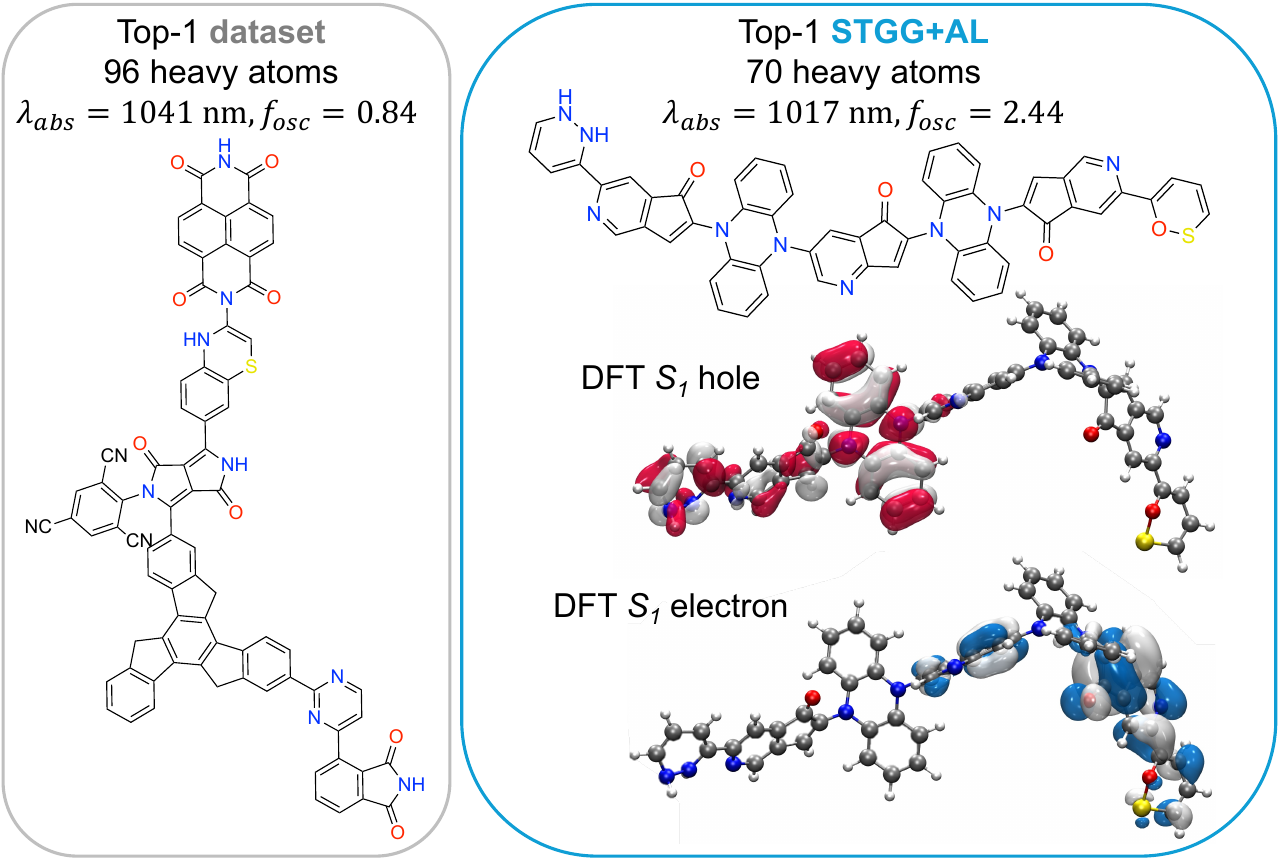}
    \caption{Case study of the top-1 molecule with NIR absorption but the highest $f_\text{osc}$.}
    \label{fig:ir_fosc}
    %\phantomsubcaption\label{fig:data_best2}
    %\phantomsubcaption\label{fig:fosc}
    \end{figure}

\begin{figure}[htbp]
    \centering
    \begin{minipage}{0.4\textwidth} 
        \centering 
        \includegraphics[width=\linewidth]{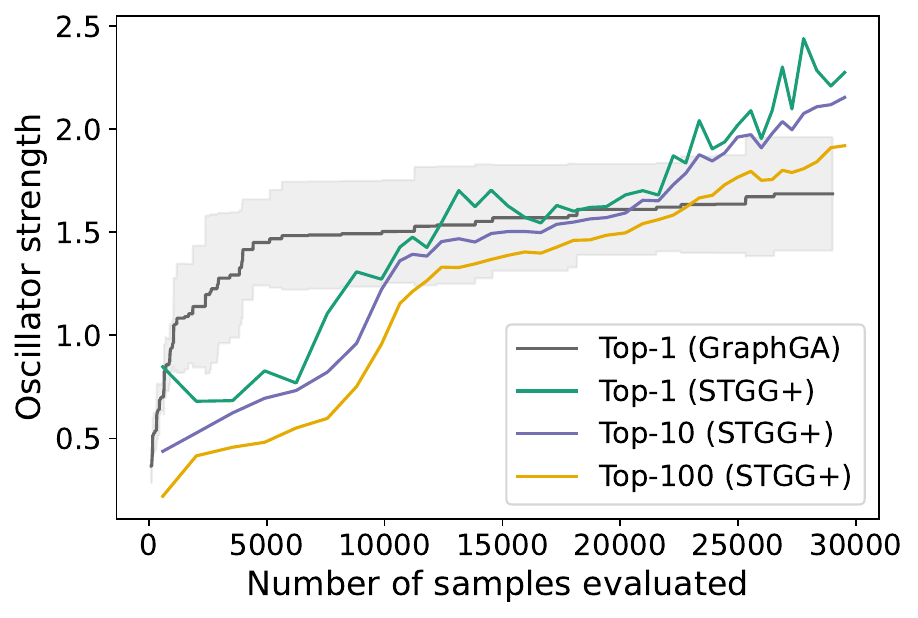}
    \end{minipage}%\hfill 
    \begin{minipage}{0.4\textwidth} 
        \centering
        \includegraphics[width=\linewidth]{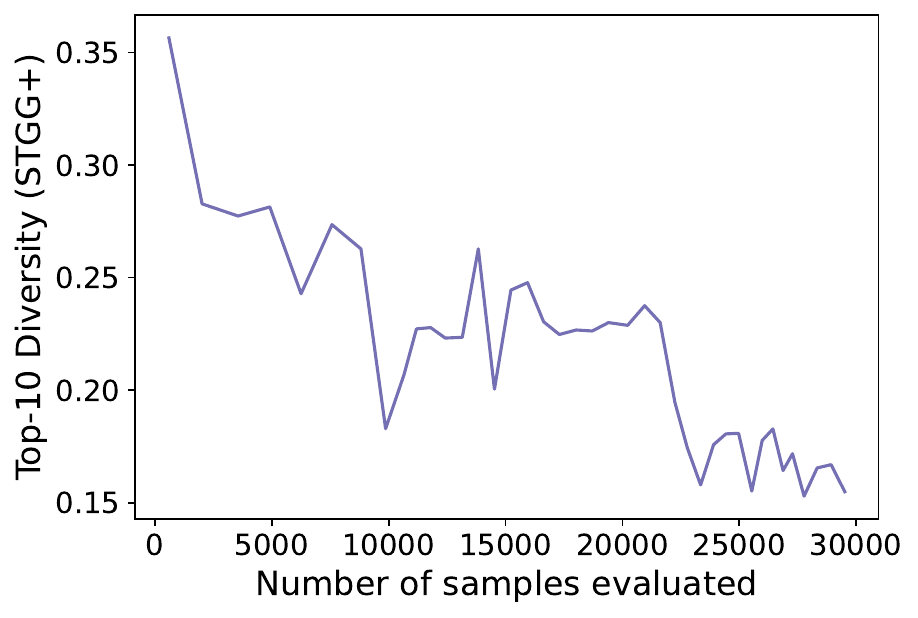} 
    \end{minipage}
    \caption{Maximizing $f_\text{osc}$ using active learning with constraints: near-IR absorption ($\lambda_{abs} \ge 1000$ nm), max 70 heavy atoms, max ring-size of 6. STGG+ (top-1, top-10, top-100; from a single run) vs GraphGA (top-1; average and 95\% confidence interval over 3 runs).}
    \label{fig:stgg_fosc2}
\end{figure}

\begin{table}[htbp]
    \caption{Comparing STGG+AL to current molecular design baselines}
    \centering
    \begin{tabular}{l|c|c}
    \hline
     Method & Max $f_\text{osc}$ & Oracle calls \\
    & mean (standard-deviation) & mean (standard-deviation) \\
    \hline
    \multicolumn{3}{c}{Maximizing $f_\text{osc}$} \\
    \hline
     Dataset (no atoms limit) & 9.30 & 0 \\
     \hline
     STGG+  & 13.01 & 0 \\
     STGG+ with active learning & \textbf{27.69} & 30.0K \\
     REINVENT4 & \phantom{1}4.53 (0.17) & 30.0K \\
     GraphGA & 14.56 (1.84) & 29.6K (4.0K) \\
    \hline
    \multicolumn{3}{c}{Maximizing $f_\text{osc}$ in short-wave infrared range} \\
    \hline
     Dataset (no atoms limit) & 0.84 & 0 \\
     \hline
     STGG+  & 0.85 & 0 \\
     STGG+ with active learning  & \textbf{2.44} & 30.0K \\
     REINVENT4 & 0.36 (0.03) & 30.0K \\
     GraphGA & 1.70 (0.28) & 30.9K (1.9K) \\
    \hline
\end{tabular}
\label{table:maintable}
\end{table}

\subsection{Summary}

From Figures \ref{fig:stgg_fosc1} and \ref{fig:stgg_fosc2}, we see that STGG+ already generates high $f_\text{osc}$ molecules (similar to or above the best value found in the Conjugated-xTB dataset) before starting the active learning, which shows strong out-of-distribution generalization. Over the active learning duration, it learns to generate even high $f_\text{osc}$molecules by actively expanding its known region. STGG+ improves consistently over time. Meanwhile, GraphGA learns quickly, but then it saturates to a local optimum and cannot improve further. We note that the molecules produced by the baseline methods have undesirable chemical features such as exotic functional groups (e.g. carbonofluoridoimidic acid) or non-conjugated components (e.g. tetraalkylammonium salt), as shown in Appendix \ref{app:4}. 

The final results for both tasks are shown in Table \ref{table:maintable}. STGG+ with active learning obtains the molecules with the highest $f_\text{osc}$. REIVENT4 performs poorly, while GraphGA manages to obtain high $f_\text{osc}$ molecules (albeit at a much lower value than STGG+).

\section{Conclusion}

STGG+ has strong out-of-distribution capabilities up to a certain limit. Active learning allows it to generate more out-of-distribution molecules with improved properties by expanding its realm of knowledge over time. We find that STGG+ with active learning is more sample-efficient compared to RL methods in search of $\pi$-conjugated molecules with high $f_\text{osc}$ and NIR dyes as simulated by semiempirical quantum chemistry methods. The generated molecules are computationally validated and rationalized using the more accurate TD-DFT methods. % Furthermore, while RL methods can generate high-reward molecules, the resulting molecules may not respect proper valency rules in molecules. Meanwhile, STGG+ ensures that molecules have proper valency by design.

We plan to expand our approach to model more complex optoelectronic properties such as fluorescence, which currently remains computationally cost-prohibitive to screen. We note that our method is not without limitation; for example, reward models such as sTDA-xTB (or TD-DFT) often fail to accurately reflect experimental results when pushed to the boundaries, and the diversity of discovered scaffolds can be further improved. 

\clearpage

\bibliographystyle{plainnat}
\bibliography{biblio}

\newpage
\appendix
\section{Appendix}

\subsection{Architecture and training details}\label{app:1}

Our STGG+ model uses mainly the same settings as \citet{stgg+} with some exceptions. 

The model is a 3-layer transformer with 16 attention heads, SwiGLU \citep{hendrycks2016gaussian, shazeer2020glu} with expansion scale of 2, no bias term \citep{chowdhery2023palm}, Flash Attention \citep{dao2022flashattention, dao2023flashattention}, RMSNorm \citep{zhang2019root}, Rotary embeddings \citep{su2024roformer}, residual-path weight initialization \citep{radford2019language}. 

We use min-max normalization for pre-processing the properties. We train the model using AdamW \citep{adam, adamw} with $\beta_1=0.9$, $\beta_2=0.95$, and weight decay 0.1. Since the molecules are large, we use a batch size of 128, learning rate of $2.5e-4$, and max sequence length of 700. 

\citet{stgg+} trained for 50 epochs on Zinc \citep{sterling2015zinc} which has 250K molecules; this amounts to 12.5M total training samples seen. Since xTB has around 2.9M molecules, we pre-train for 5 epochs in order to process a similar amount of training samples (14.5M).

Fine-tuning is done for 100 epochs at whichever number of molecules is available ($\leq$ 2000 since we generate 2000 molecules before applying the constraints). This is effectively equivalent to training on $\leq$ 200K samples, which is around 1.4\% if the pre-training time. Given our 40 active learning steps, around 55\% of the training time that is spent on fine-tuning.

For generation of molecules, we sample uniformly from a range of hyperparameters in order to get diversity. While \citet{stgg+} only does this for the guidance term in the classifier-free guidance \citep{ho2022classifier}, we also do it for temperature. We sample a temperature uniformly between 0.6 and 1.1 and guidance between 0.5 and 1.5. This range of values as not been tuned so its possible that there are better choices. We always sample 2000 molecules and remove duplicates and those not respecting the given constraints (max ring size of 6, maximum of 70 atoms).

For the RL baselines, we scaled the oscillator strength to become a reward between 0 and 1 in the following way: $R= min(max(f_{\text{osc}}/27.70,0),1)$ for task 1 and $R= min(max(f_{\text{osc}}/2.44,0),1)$ for the task 2. 27.7 and 2.44 are respectively the maximum $f_{\text{osc}}$ obtained in task 1 and 2 by STGG+. None of the RL baselines reached 1.0 or above (otherwise, we would have rescaled them differently). We also enforced the ring size maximum of 6, number of heavy atoms $\leq$ 70, and NIR range by setting the reward to 0 when any of these constraints are violated.

\subsection{RL baseline details}\label{app:rl}

For REINVENT4, we use a standard config as provided by the authors with minimal modifications: using the reinvent prior and agent, batch-size=100, remove duplicates, randomizing smiles, a maximum number of steps of 300 (to reach around 30K Oracle calls), using the default DAP with sigma=128 and learning rate 0.0001, the default diversity filter (Identical Murcko Scaffold), and the default unwanted SMARTS penalty (to penalized wonky molecules). 

For GraphGA, to maximize performance, we followed the good choice of hyperparameters by \citep{tripp2023genetic} which consist in only generating 5 offspring by generation, but generating as many generation as desired (in our case 7500 generations to reach around 30K Oracle calls).

To give an head-start to the RL baselines, we fed the top-100 molecules for each task from the Conjugated-xTB dataset as prior molecules. For REINVENT4, at each iteration, 10 of these 100 molecules where randomly picked and added to the mini-batch to aid learning. For GraphGA these top-100 molecules formed the initial population.

\clearpage
\subsection{Architecture diagram}\label{app:2}

\begin{figure}[htb]
    \centering
\includegraphics[width=0.65\linewidth]{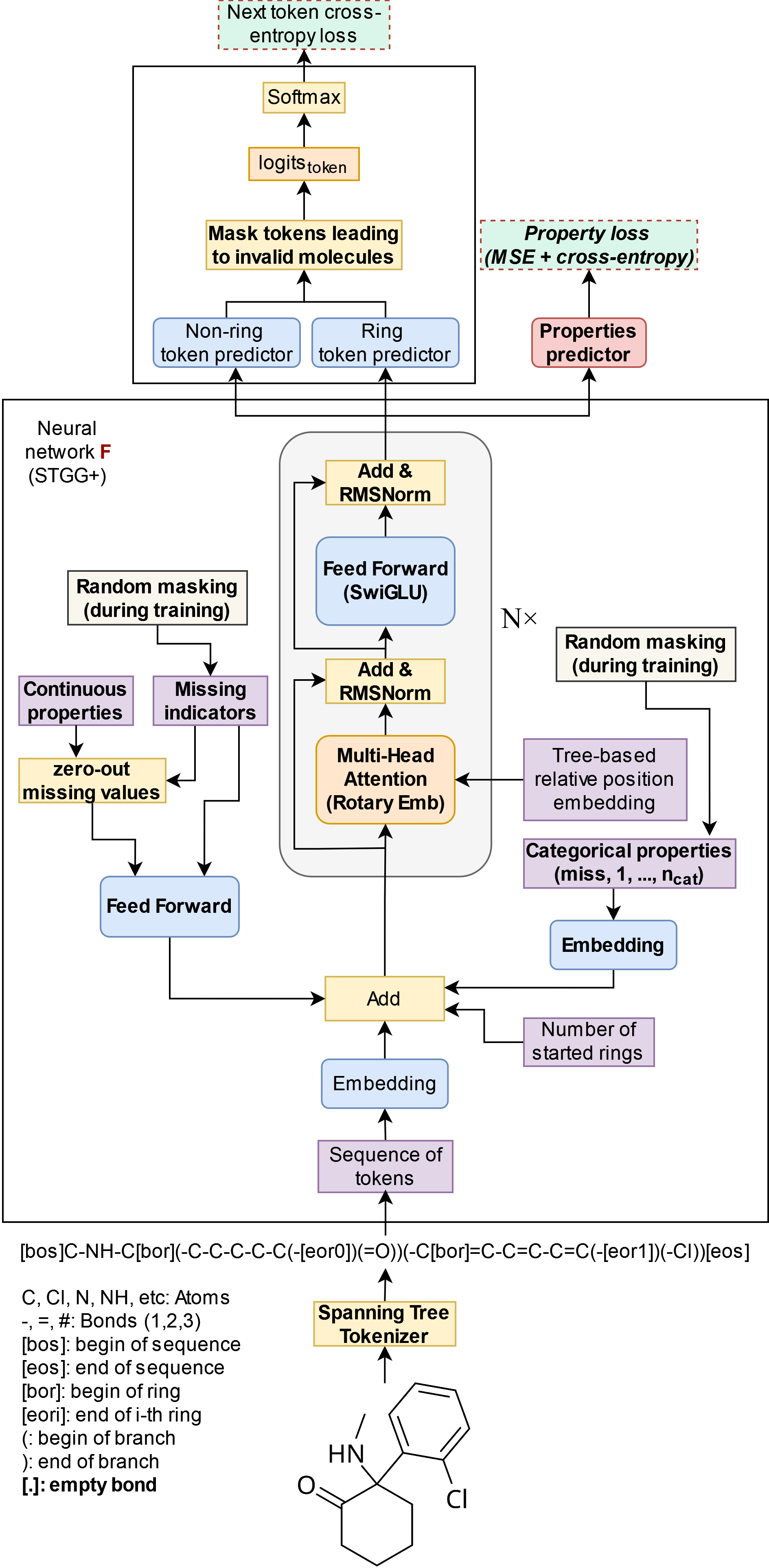}
    \caption{STGG+ architecture. The molecule is tokenized and embedded. The number of started rings and embeddings of continuous and categorical properties are added, and the output is passed to the Transformer. The Transformer output is then split to produce 1) the predicted property and 2) the token predictions (masked to prevent invalid tokens). Novel components compared to STGG~\citep{ahn2021spanning} are in \textbf{bold}. The figure was taken from \citet{stgg+}.}
    \label{fig:STGG1}
\end{figure}

The architecture of STGG+ is shown in Figure \ref{fig:STGG1}.

\clearpage
\subsection{The best SMILES generated by STGG+}\label{app:3}

\begin{table}[ht]
    \setlength{\tabcolsep}{1pt}
    \caption{Top-10 molecules found using STGG+ with active-learning}
    \centering
    \scriptsize
    \begin{tabular}{c|c|c}
    \hline
     $f_\text{osc}$ & Similarity & SMILES \\
    \hline
    \multicolumn{3}{c}{maximizing $f_\text{osc}$} \\
    \hline
27.69 & 0.88 & \text{Cc1cc2oc3cc(C\#Cc4cc5[nH]c(-c6cc7sc(-c8cc9sc(-c\%10cc\%11sc(-c\%12cc\%13sc(C\#CC\%14=C\%15N=CC=[N+]} \\

& & \text{\%15[B-](Br)(Br)n\%15cccc\%15\%14)cc\%13[nH]\%12)cc\%11[nH]\%10)cc9s8)cc7s6)cc5s4)sc3c2s1} \\
27.66 & 0.88 & \text{Brc1cc2c(s1)-c1sc(C\#Cc3cc4[nH]c(-c5cc6sc(-c7cc8sc(-c9cc\%10sc(-c\%11cc\%12sc(C\#CC\%13=C\%14N=CC=[N+]} \\

& & \text{\%14[B-](Br)(Br)n\%14cccc\%14\%13)cc\%12[nH]\%11)cc\%10[nH]9)cc8s7)cc6s5)cc4s3)cc1C2} \\
27.55 & 0.87 & \text{Clc1cc2[nH]c3cc(C\#Cc4cc5sc(-c6cc7sc(-c8cc9sc(-c\%10cc\%11sc(-c\%12cc\%13sc(C\#CC\%14=C\%15N=CC=[N+]} \\

& & \text{\%15[B-](Br)(Br)n\%15cccc\%15\%14)cc\%13[nH]\%12)cc\%11[nH]\%10)cc9s8)cc7s6)cc5s4)sc3c2s1} \\
27.50 & 0.85 & \text{Cc1cc2[nH]c3cc(C\#Cc4cc5[nH]c(-c6cc7sc(-c8cc9sc(-c\%10cc\%11sc(-c\%12cc\%13sc(C\#CC\%14=C\%15N=CC=[N+]} \\

& & \text{\%15[B-](Br)(Br)n\%15cccc\%15\%14)cc\%13[nH]\%12)cc\%11[nH]\%10)cc9s8)cc7s6)cc5s4)sc3c2s1} \\
27.38 & 0.89 & \text{Brc1cc2c(s1)-c1sc(C\#Cc3cc4[nH]c(-c5cc6sc(-c7cc8sc(-c9cc\%10sc(-c\%11cc\%12[nH]c(C\#CC\%13=C\%14N=CC=[N+]} \\

& & \text{\%14[B-](Br)(Br)n\%14cccc\%14\%13)cc\%12s\%11)cc\%10[nH]9)cc8s7)cc6s5)cc4s3)cc1C2} \\
27.35 & 0.86 & \text{Clc1cc2[nH]c3cc(C\#Cc4cc5[nH]c(-c6cc7sc(-c8cc9sc(-c\%10cc\%11sc(-c\%12cc\%13sc(C\#CC\%14=C\%15N=CC=[N+]} \\

& & \text{\%15[B-](Br)(Br)n\%15cccc\%15\%14)cc\%13[nH]\%12)cc\%11[nH]\%10)cc9s8)cc7s6)cc5s4)sc3c2s1} \\
27.35 & 0.88 & \text{Brc1cc2sc3cc(C\#Cc4cc5[nH]c(-c6cc7sc(-c8cc9sc(-c\%10cc\%11sc(-c\%12cc\%13[nH]c(C\#CC\%14=C\%15N=CC=[N+]} \\

& & \text{\%15[B-](Br)(Br)n\%15cccc\%15\%14)cc\%13s\%12)cc\%11[nH]\%10)cc9s8)cc7s6)cc5s4)[nH]c3c2s1} \\
27.27 & 0.87 & \text{Clc1cc2[nH]c3cc(C\#Cc4cc5[nH]c(-c6cc7sc(-c8cc9sc(-c\%10cc\%11[nH]c(-c\%12cc\%13sc(C\#CC\%14=C\%15N=CC=[N+]} \\

& & \text{\%15[B-](Br)(Br)n\%15cccc\%15\%14)cc\%13[nH]\%12)cc\%11s\%10)cc9s8)cc7s6)cc5s4)sc3c2s1} \\
27.22 & 0.92 & \text{Cc1cc2c(o1)-c1[nH]c(C\#Cc3cc4[nH]c(-c5cc6sc(-c7cc8sc(-c9cc\%10sc(-c\%11cc\%12[nH]c(C\#CC\%13=C\%14N=CC=[N+]} \\

& & \text{\%14[B-](Br)(Br)n\%14cccc\%14\%13)cc\%12[nH]\%11)cc\%10[nH]9)cc8s7)cc6s5)cc4s3)cc1C2} \\
27.09 & 0.87 & \text{Br[B-]1(Br)n2cccc2C(C\#Cc2cc3[nH]c(-c4cc5[nH]c(-c6cc7sc(-c8cc9sc(-c\%10cc\%11sc(C\#Cc\%12cc\%13sc\%14cc[nH]} \\

& & \text{c\%14c\%13[nH]\%12)cc\%11[nH]\%10)cc9s8)cc7s6)cc5s4)cc3s2)=C2N=CC=[N+]21} \\
\hline
\multicolumn{3}{c}{maximizing $f_\text{osc}$ in short-wave infrared range} \\
\hline
2.44 & 0.79 & O=C1C(N2c3ccccc3N(C3=Cc4ncc(N5c6ccccc6N(C6=Cc7cnc(C8=CC=CNN8)cc7C6=O) \\
& & c6ccccc65)cc4C3=O)c3ccccc32)=Cc2cnc(C3=CC=CSO3)cc21 \\
2.30 & 0.74 & O=C1C(N2c3ccccc3N(C3=Cc4cnc(N5c6ccccc6N(C6=Cc7cnc(C8=CC=CSN8)cc7C6=O) \\
& & c6ccccc65)cc4C3=O)c3ccccc32)=Cc2cnc(C3=CC=CNN3)cc21 \\
2.28 & 0.75 & O=C1C(N2c3ccccc3N(C3=Cc4cnc(C5=CC=CSN5)cc4C3=O) \\
& & c3ccccc32)=Cc2ccc(N3c4ccccc4N(C4=Cc5cc(C6=CC=CNN6)ncc5C4=O)c4ccccc43)cc21 \\
2.27 & 0.81 & O=C1C(N2c3ccccc3N(C3=Cc4cc(N5c6ccccc6N(C6=Cc7cnc(C8=CC=CSN8)nc7C6=O) \\
& & c6ccccc65)cnc4C3=O)c3ccccc32)=Cc2cnc(C3=CC=CNO3)cc21 \\
2.22 & 0.83 & O=C1C(N2c3ccccc3N(c3cnc4c(c3)C(=O)C(N3c5ccccc5N(C5=Cc6ccc(C7=CC=CSO7)nc6C5=O) \\
& & c5ccccc53)=C4)c3ccccc32)=Cc2cnc(C3=CC=CSN3)cc21 \\
2.21 & 0.80 & O=C1C(N2c3ccccc3N(C3=Cc4ccc(C5=CC=CSN5)nc4C3=O)c3ccccc32) \\
& & =Cc2ncc(N3c4ccccc4N(C4=Cc5cnc(C6=CC=CSN6)nc5C4=O)c4ccccc43)cc21 \\
2.20 & 0.79 & O=C1C(N2c3ccccc3N(C3=Cc4cc(N5c6ccccc6N(C6=Cc7cnc(C8=CC=CSO8) \\
& & cc7C6=O)c6ccccc65)cnc4C3=O)c3ccccc32)=Cc2cnc(C3=CC=CNS3)cc21 \\
2.21 & 0.81 & O=C1C(N2c3ccccc3N(C3=Cc4cc(N5c6ccccc6N(C6=Cc7cnc(C8=CC=CNN8) \\
& & nc7C6=O)c6ccccc65)cnc4C3=O)c3ccccc32)=Cc2cnc(C3=CC=CSN3)cc21 \\
2.19 & 0.83 & O=C1C(N2c3ccccc3N(c3ccc4c(n3)C=C(N3c5ccccc5N(C5=Cc6cnc(C7=CC=CNN7)nc6C5=O) \\
& & c5ccccc53)C4=O)c3ccccc32)=Cc2cnc(C3=CC=CSN3)cc21 \\
2.18 & 0.78 & O=C1C(N2c3ccccc3N(C3=Cc4cnc(C5=CC=CNN5)nc4C3=O)c3ccccc32)=Cc2ccc \\
& & (N3c4ccccc4N(C4=Cc5ncc(C6=CC=CNN6)nc5C4=O)c4ccccc43)cc21 \\
\hline
\end{tabular}
\end{table}

\clearpage
\subsection{The best molecules generated by baseline methods}\label{app:4}

\subsubsection{STGG+}

\begin{figure}[h]
    \centering 
    \includegraphics[width=1\linewidth]{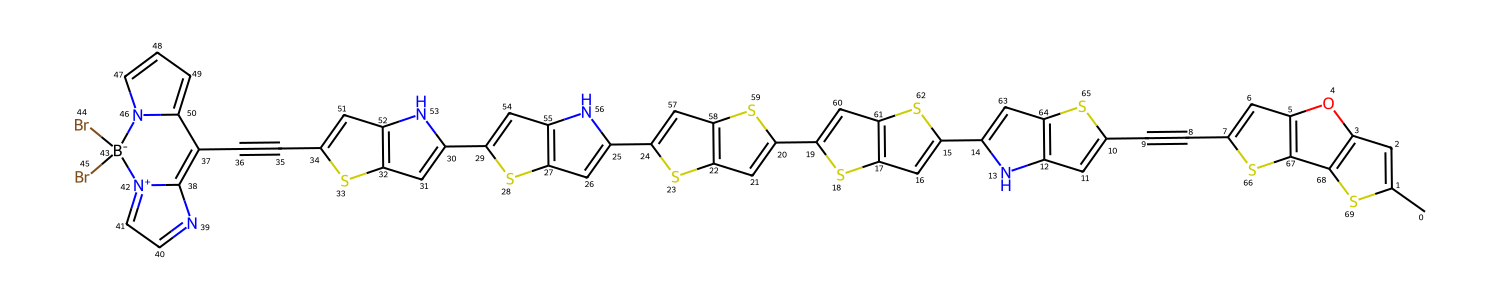}
    \caption{STGG+ Top-1 molecule with the highest $f_\text{osc}$ out of a single run ($f_\text{osc}=27.69$).}
    \label{fig:beststgg+}
\end{figure}

\begin{figure}[h]
    \centering 
    \includegraphics[width=1\linewidth]{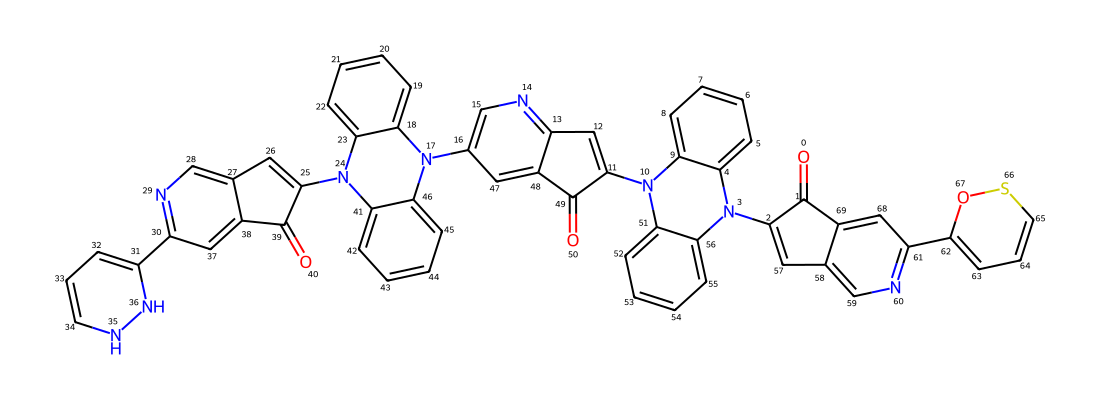}
    \caption{STGG+ Top-1 molecule with the highest $f_\text{osc}$ and near-IR absorption out of a single run ($f_\text{osc}=2.44$).}
    \label{fig:beststgg+_ir}
\end{figure}

The top-1 molecules generated by STGG+ are shown in Figures \ref{fig:beststgg+}-\ref{fig:beststgg+_ir}. The molecules are generally sensible and plausible, respecting proper valency. We note that certain structures can still be exotic (e.g. 1,2-oxathiine, dipyrromethene borondibromide), but they are nevertheless previously known compounds and do not affect the core scaffolds.

\clearpage
\subsubsection{REINVENT4}

\begin{figure}[h]
    \centering 
    \includegraphics[width=1\linewidth]{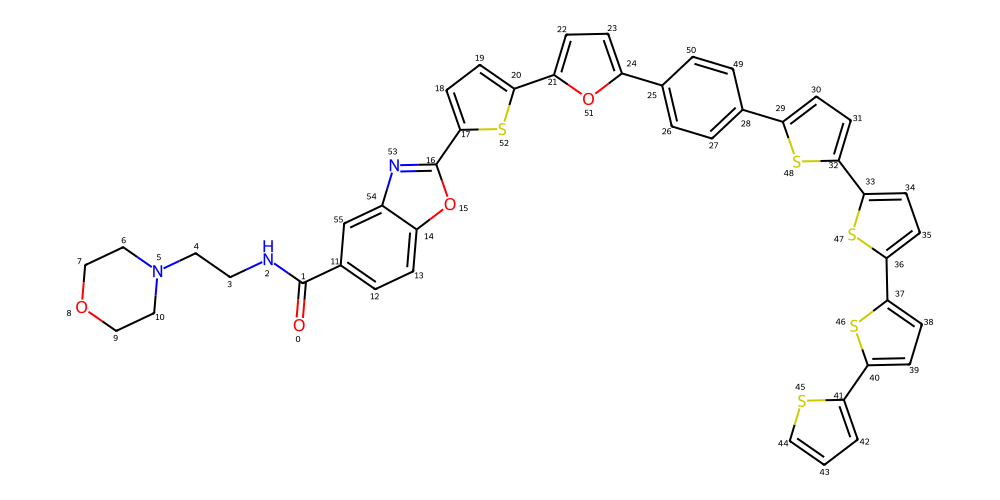}
    \caption{REINVENT4 Top-1 molecule with the highest $f_\text{osc}$ out of 3 runs ($f_\text{osc}=4.65$). This polythiophene derivative has a long non-conjugated group that does not contribute to $f_\text{osc}$.}
    \label{fig:bestRE}
\end{figure}

\begin{figure}[h]
    \centering 
    \includegraphics[width=1\linewidth]{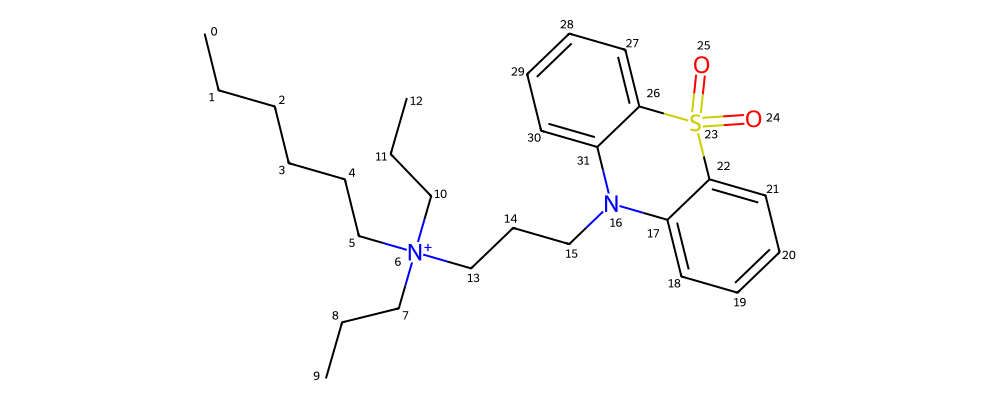}
    \caption{REINVENT4 Top-1 molecule with the highest $f_\text{osc}$ and near-IR absorption out of 3 runs ($f_\text{osc}=0.40$). This phenothiazine dioxide lacks conjugation, is not unexpected to be absorptive in NIR, and has an non-conjugated tetralkylammonium salt pendant group.}
    \label{fig:bestRE_ir}
\end{figure}

The top-1 molecules generated by REINVENT4 are shown in Figures \ref{fig:bestRE}-\ref{fig:bestRE_ir}. 

\clearpage
\subsubsection{GraphGA}

\begin{figure}[h]
    \centering \includegraphics[width=1\linewidth]{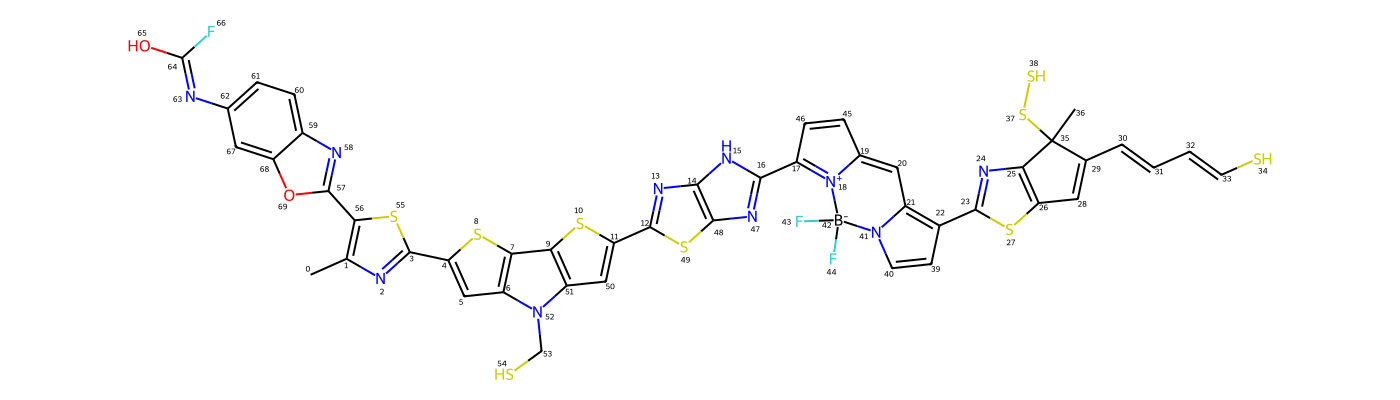}
    \caption{GraphGA Top-1 molecule with the highest $f_\text{osc}$ out of 3 runs ($f_\text{osc}=15.81$). This molecule has several undesirable functional groups including carbonofluoridoimidic acid and disulfaneylmethylcyclopenta[d]thiazole.}
    \label{fig:bestGraphGA}
\end{figure}

\begin{figure}[h]
    \centering 
    \includegraphics[width=1\linewidth]{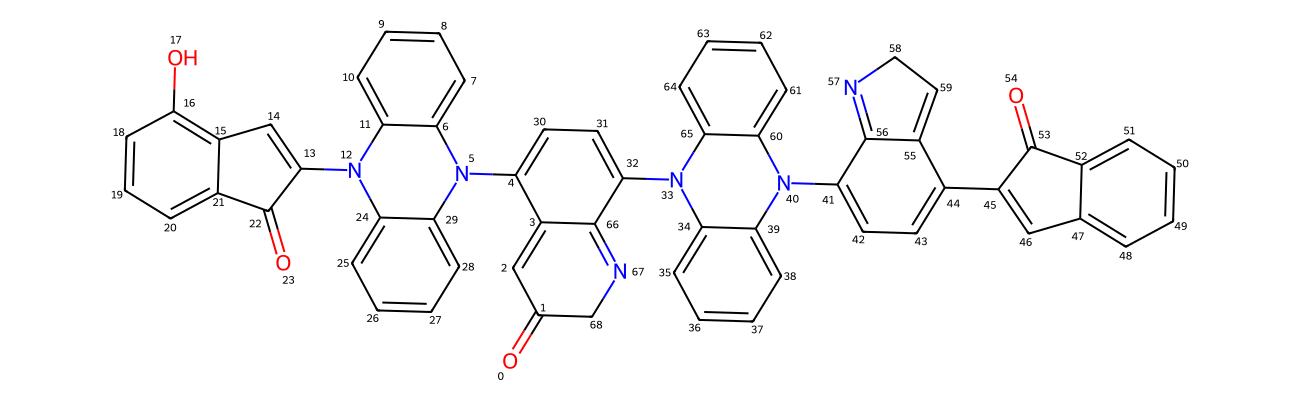}
    \caption{GraphGA Top-1 molecule with the highest $f_\text{osc}$ and near-IR absorption out of 3 runs ($f_\text{osc}=2.03$). This molecule is largely chemically sound, with the exception of the quinolin-3(2H)-one.}
    \label{fig:bestGraphGA_ir}
\end{figure}

The top-1 molecules generated by GraphGA are shown in Figures \ref{fig:bestGraphGA}-\ref{fig:bestGraphGA_ir}. The best molecule for maximizing $f_\text{osc}$ without constraint is extremely implausible and unlikely to be synthesizable.

\end{document}